\documentclass{article}

\usepackage{arxiv}

\usepackage[utf8]{inputenc} 
\usepackage[T1]{fontenc}    
\usepackage{hyperref}       
\usepackage{url}            
\usepackage{booktabs}       
\usepackage{amsfonts}       
\usepackage{nicefrac}       
\usepackage{microtype}      
\usepackage{lipsum}

\usepackage{cite}
\usepackage[colorinlistoftodos]{todonotes}
\usepackage{floatrow}
\usepackage{float}
\usepackage[caption=false,font=normalsize,labelfont=sf,textfont=sf]{subfig}

\title{Chest Area Segmentation in Depth Images of Sleeping Patients}

\author{Yoav~Goldstein,~\IEEEmembership{Member,~IEEE,}
        Martin~Sch\"{a}tz,~\IEEEmembership{Fellow,~OSA,}
        and~Mireille~Avigal,~\IEEEmembership{Life~Fellow,~IEEE}
\thanks{Y. Goldstein and M. Avigal was with the Department of Mathematics and Computer Science, Open University of Israel, Raanana, Israel}
\thanks{M. Sch\"{a}tz. with University of Chemistry and Technology in Prague, Department of Computing and Control Engineering, Czech Republic.}
\thanks{Manuscript received April 19, 2005; revised August 26, 2015.}}

\author{
  Yoav~Goldstein\\
  Department of Mathematics and Computer Science\\
  Open University of Israel
  Raanana, Israel\\
   \And
    Martin~Sch\"{a}tz \\
  University of Chemistry and Technology in Prague\\
  Department of Computing and Control Engineering\\
  Czech Republic\\
  \AND
   Mireille~Avigal\\
  Department of Mathematics and Computer Science\\
  Open University of Israel
  Raanana, Israel\\
}

\begin{document}
\maketitle

\begin{abstract}
Although the field of sleep study has greatly developed over the recent years, the most common and efficient way to detect sleep issues remains a sleep examination performed in a sleep laboratory, in a procedure called Polysomnography (PSG). This examination measures several vital signals during a full night's sleep using multiple sensors connected to the patient's body. Yet, despite being the golden standard, the connection of the sensors and the unfamiliar environment inevitably impact the quality of the patient's sleep and the examination itself. 

Therefore, with the novel development of more accurate and affordable 3D sensing devices, new approaches for non-contact sleep study emerged. These methods utilize different techniques with the purpose to extract the same sleep parameters, but remotely, eliminating the need of any physical connections to the patient's body. However, in order to enable reliable remote extraction, these methods require accurate identification of the basic Region of Interest (ROI) i.e. the chest area of the patient, a task that is currently holding back the development process, as it is performed manually for each patient. 

In this study, we propose an automatic chest area segmentation algorithm, that given an input set of 3D frames of a sleeping patient, outputs a segmentation image with the pixels that correspond to the chest area, and can then be used as an input to subsequent sleep analysis algorithms. Except for significantly speeding up the development process of the non-contact methods, accurate automatic segmentation can also enable a more precise feature extraction and it is shown it is already improving sensitivity of prior solutions on average 46.9\% better compared to manual ROI selection. All mentioned will place the extraction algorithms of the non-contact methods as a leading candidate to replace the existing traditional methods used today.
\vspace{1cm}
\end{abstract}


\section{Introduction}
{A} {sleeping}  disorder, or Somnipathy, is a medical disorder of the sleep patterns, which can be serious enough to affect one's physical and in some cases even physiological state
\cite{Prochazka2016,Gibson2004,Al-Naji2017,Flemons2003,Palmero2017,Schatz2020}.
There are many known and unknown types of sleep disorders as well as many methods to discover and diagnose them, with the most familiar and commonly used being a sleep study performed in a sleep laboratory. This method, called Polysomnography (PSG) \cite{Hirshkowitz2014a}, is an examination that combines a measurement of several vital signals, such as Heart Rate (HR), Electroencephalography (EEG) and respiratory rate, using multiple sensors connected to the patient's body during a full night's sleep. PSG is performed in sleep laboratories, which generally exist in large hospitals, where there is also a large number of sleep experts and the necessary sleep study equipment, capable of detecting various types of sleep disorders. 

Despite the general thought, sleep disorders are very common among the population. For example, Obstructive Sleep Apnea (OSA) caused by periodic obstruction of the upper airways during the sleep can be diagnosed among 5-15\% of the general population \cite{Young2002}. However, the existing method of sleeping a whole night in a clinic in the hospital does not encourage the majority of the population to go through a diagnosis. Moreover, the multiple sensors connected to the patient's body directly affect the quality and characteristics of his or her sleep.

Therefore, in recent years, a new kind of sleep monitoring methods began to emerge - non-contact methods. These methods study the sleep process requiring no physical connection of cables and sensors to the patient's body \cite{Dafna2015,Harte2016}. Instead, they utilize a variety of technologies, such as radars \cite{Kagawa2013}, audio recordings \cite{Yang2017} and depth sensors \cite{Schatz2015,Prochazka2016,Nguyen2016,Al-Naji2017,Yang2017,SR300perf, Schatz2020}.
The development of such methods will enable patients to examine their sleep in their home environment instead of in a sleep clinic, thus both increasing diagnosis rates as well as achieving a measurement that better represents the actual sleep of the patient. 


\subsection{The Breathing Area}
Many of the current sleep study methods require computation of the breathing frequency of the patient as a starting point, from which other sleep features such as sleep stages and sleep regularity can be deduced \cite{Prochazka2016}. In order to extract the breathing frequency, an input in the form of the current position of the person or the location of its chest area is required. To achieve that, there are many new methods for body feature recognition using depth data \cite{Schwarz2012,Garn2016,Chen2013} and many algorithms are used to evaluate the pose of the body \cite{Schwarz2011}. However, the available algorithms do not address the scenario of sleeping patients, but rather handle chest detection in other body poses such as standing and walking \cite{Harte2016}. Furthermore, these algorithms are not optimized for sleep feature extraction purposes. In this research, we handle sleeping poses of patients as captured during their sleep and focus on the detection of the chest movement in their depth video streams.

The chest area is the area in the image in which the breathing can be seen both locally in the image and also through time in the video, and hence detecting this region allows an algorithm that requires the breathing area as an input, such as the algorithm described in  \cite{Prochazka2016} to infer the basic signals for the sleep feature extraction. In this research, we will define a pixel in the chest area as a pixel in the image that is influenced by the rib cage movement. Most of the time, these pixels will indeed be physically located on the patient's chest. However, in other cases, for example when the chest is not directly exposed to the camera, the movement can be detected in other physical areas as well.


\subsection{Sleeping Patients Depth Dataset} 
In the study described in \cite{Prochazka2016}, the sleep feature extraction algorithm used videos recorded by depth camera during a sleep examination. 
As the input for the segmentation algorithm in this research are 3D frames, each input is essentially a depth map of the patient and the scene around him at that moment in time. A depth map, like any regular RGB image is a numerical matrix containing values corresponding to the amount of light that is captured by the different pixels, but unlike a regular 2D image, it also contains another dimension that captures the distance from the camera for every pixel.




The sleeping patients' database was acquired in a sleep clinic located in the faculty hospital Hradec Králové, Czech republic using a Microsoft Kinect depth Camera as a 3D video stream. 
In addition, in order to compare the performance of the method in \cite{Prochazka2016} to the gold standard, the depth stream was captured in parallel to a PSG examination.
A frame from the video can be seen in Fig.~\ref{fig:examplePatientFigure}. This Figure captures a sleeping patient from the side, in a posture as can be seen in Fig.~\ref{fig:examplePatientFigure}:

\begin{figure}[!htb]
    \centering
    \subfloat[the depth map]{{\includegraphics[width=6cm]{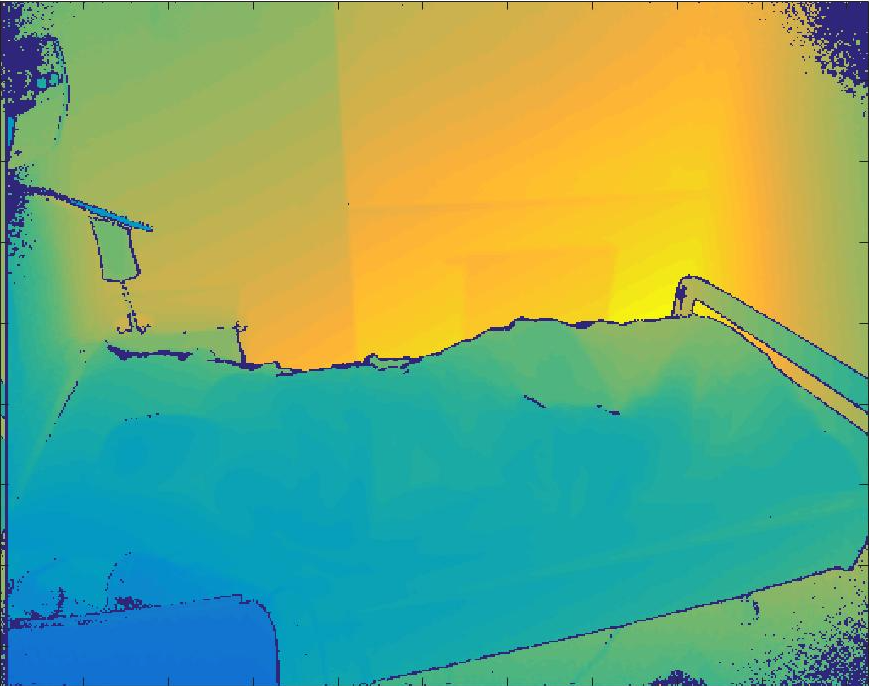} }}%
    \qquad
    \subfloat[camera direction]{{\includegraphics[width=6cm]{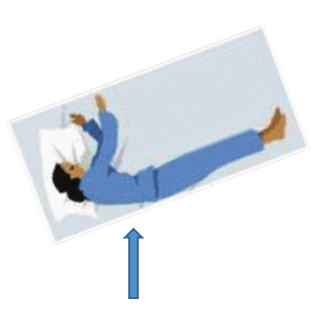} }}%
    \caption{An example of a sleeping patient depth map (a) with the camera direction (b)}%
    \label{fig:examplePatientFigure}%
\end{figure}


\section{Methods and Methodologies}
\begin{figure}
[!htb]
\begin{center}
\includegraphics{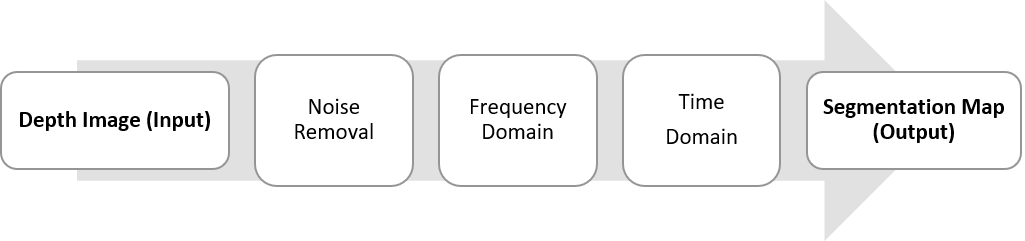}
\end{center}
\caption{\small \sl A system overview diagram of the three main stages of the segmentation algorithm.
\label{fig:SystemOverviewDiagram}}  
\end{figure}

The implementation of the algorithm consisted of three main stages, as can be observed on
the system overview diagram in Fig  \ref{fig:SystemOverviewDiagram}.

\subsection{Noise Removal}
In order to detect the relevant pixels that belong to the chest area, to allow algorithms such as in \cite{Prochazka2016} to extract the breathing frequency more accurately, the following were implemented:

\subsubsection{Basic Threshold and Median Filtering}
As explained, depth pixels encapsulates the distance value of a pixel from the camera, ranging the value of each pixel between 0 and the maximal distance the camera can measure (usually in millimetres). However, in a depth camera image output, a zero value pixel does not indicate a pixel with zero distance but a pixel which its depth value cannot be determined, which translates to a black pixel in the visualized depth map. Because of the above, each of the depth maps were normalized to the values of [0, 1] using a distance threshold, which pixels valued higher then it were discarded. The threshold value was chosen to be about 3m which is close to the maximal depth distance of the cameras dynamic range. This also assures that all parts of the patient will remain in the frame.

\begin{figure}[!htb]
\begin{center}
\includegraphics[width = 0.5\linewidth]{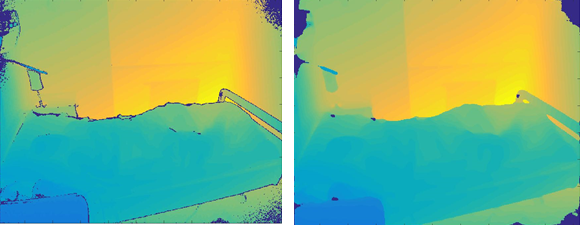}
\end{center}
\caption{\small \sl Depth Images before (left) and after (right) ‘Pepper’ noise removal by median filtering.\label{fig:PepperRemove}}  
\end{figure}

The zeroed values in the depth images needed to be removed as well. These values in the depth image resemble the ‘Salt and Pepper’ noise model, but without the white ‘Salt’ pixels. The method that was implemented to remove those ‘Pepper’ pixels is a median filter that estimates the values of those missing pixels based on the median value of their neighbours, in $10 \times 10$ radius, meaning that every pixel is replaced with the median value in a filter of radius $10 \times 10$ pixels. An example of the ‘Pepper’ noise from depth pixels with no data can be seen in Fig.~\ref{fig:PepperRemove}

\subsubsection{Boundary Frame Removal }
Depth images are characterized by a noise model that increases its intensity proportionally to the distance of a pixel to the center of the frame. In addition, as farther as a pixel from the center of the frame, higher the probability that the pixel depth value could not be determined (which results to a black pixel).
To reduce this kind of noise, margins of the image in 50 pixel width were ignored. This operation was done by setting the pixels in the margins to zero, and later ignoring them in further segmentation pixel calculations \cite{Mallick2014}. The value of 50 pixels was selected based on the location of the camera in the data that was recorded, and might change based on the characteristics of the depth sensor, if the latter will change in the future.
An example of an image after this process of margin removal can be observed in Fig.~\ref{fig:FinalNoiseRemove}a.


\subsubsection{Edges Processing}

\begin{figure}[!htb]

    \centering
    \subfloat[Margin removed image of 50 Pixels from each side]{{\includegraphics[width=4.5cm]{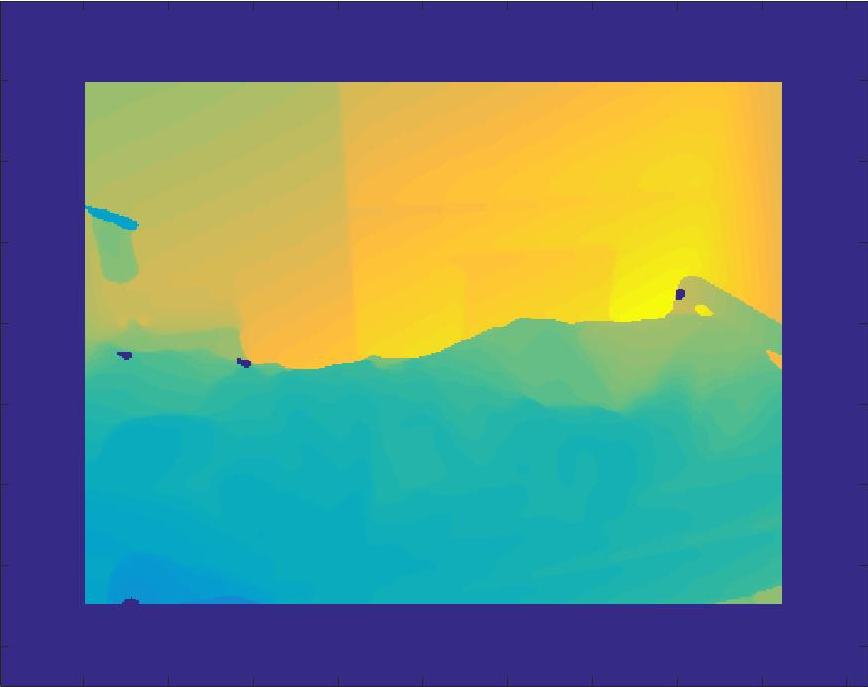} }}%
    \qquad
    \subfloat[Canny Edge map of the depth image]{{\includegraphics[width=4.5cm]{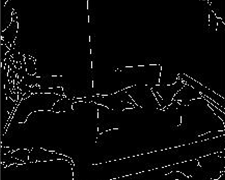} }}%
    \qquad
    \subfloat[Final noisy pixels image, allowing the segmentation algorithm to ignore noise pixels]{{\includegraphics[width=4.5cm]{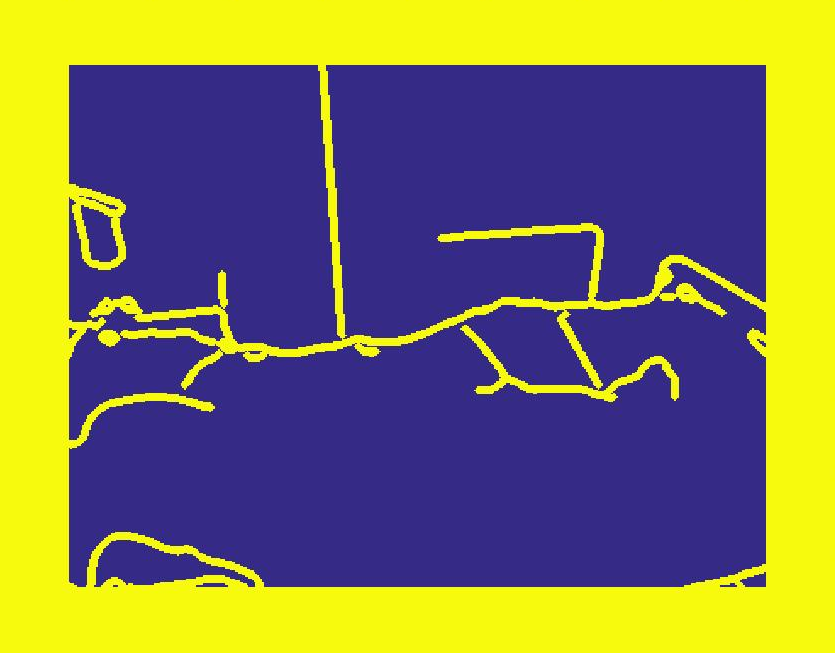} }}%
    \caption{Image with no margin (a) Canny image (b) and a binary image representing the noisy pixels to be removed (c)}%
    \label{fig:FinalNoiseRemove}%
\end{figure}

The edges in a depth image usually represents a transition from an area with a certain distance to another area with other distance range. Therefore, pixels on the edge of the depth image will be a more prone to be noise, since on each frame those pixels can count as part of one region and therefore will be assigned with a distance value from it and on another one can be assigned a value from another. Since the data processed in the research is a depth image sequence, those pixels on the edges were noisy, and needed to be ignored.
In order to locate those pixels, canny edge map was created [14]. Later on, a dilation morphological operator was applied on the image edges to better capture the edges environment. An example edge image result can be seen in Fig.~\ref{fig:FinalNoiseRemove}b.
Finally, a binary image was created, indicating in white the pixels to ignore combining all of the above, as can be seen in Fig.~\ref{fig:FinalNoiseRemove}c.

\subsection{The Segmentation Algorithm}
\subsubsection{Basic algorithmic assumptions}
The main approach to segmenting the chest area comes from the fact that it has been already proven before that depth streams of breathing patients hold the information for the extraction of breathing frequency from the chest area \cite{Prochazka2016}. Alongside this, the algorithm is developed under the following assumptions:
\begin{itemize}
\item The depth sequence must contain a sleeping patient.
\item Most of the time the patient is not moving and changing its sleeping position (except the breathing movement), and periodic posture changes.
\item There are no other moving objects in the scene apart for the patient.
\end{itemize}
It is safe to say that these assumptions are reasonable to assume because the data in this research comply with them since it captured sleeping patients during a sleep study in a sleep laboratory.

While many algorithms for moving object segmentation use some form of optical flow algorithm \cite{Schwarz2011}, in this case where the data represents distance pixels, traditional optical flow algorithm will not be effective. The reason for this is that optical flow measures the displacement of pixels in consecutive frames, and in this case, when handling depth images, an object movement may not be represented by pixel displacement. An example for this comes from the fact that movement in the tangential direction of the camera will not cause a change in the distance and value of the pixels but radial direction movement will.
However, based on the non-movement assumption, a change of value on a single-pixel can represent the changing distance of it from the camera and therefore can truly represent its movement over time.
Therefore, the main candidates for chest pixels are the pixels that change their value over time and not their location in the frame. As we assume that the main object in the frame is the sleeping patient, the only constant movement in the scene will correspond the movement of his chest while breathing (as long as we follow our second assumption and ignore some posture changes which occur less frequently over time).


\subsubsection{Frequency Domain Manipulation}
\begin{figure}[!htb]
    \centering
    \begin{floatrow}
    \subfloat[Maximal amplitude image of each depth pixel over time sequence. The bright pixels represent high amplitudes in the frequency domain.]{{\includegraphics[width=5cm]{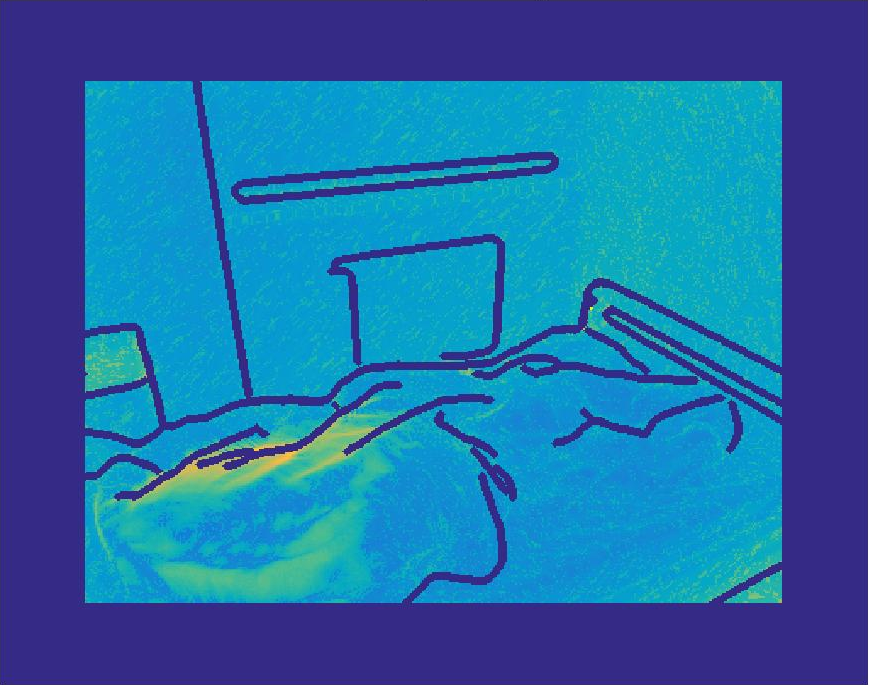} }}%
     
    \qquad
    \subfloat[Threshold on chest pixels after frequency domain, representing the main candidates for chest pixels.]{{\includegraphics[width=5cm]{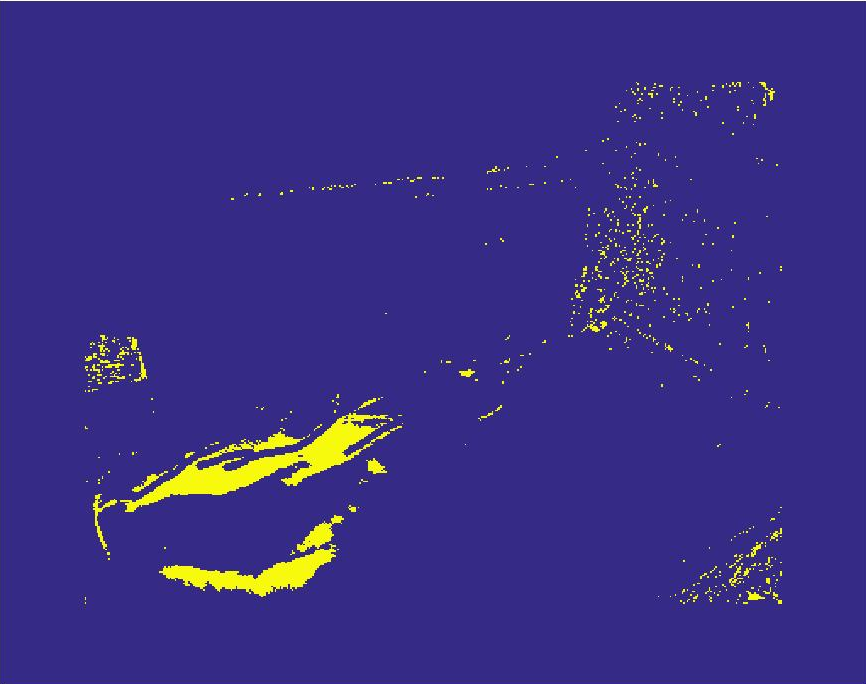} }}%
    \caption{Amplitude image and thresholded pixels}%
    \label{fig:MaximalAmplitude}%
    \end{floatrow}
\end{figure}

As shown in \cite{Prochazka2016}, the breathing frequency can be extracted from the chest area. We use this fact in order to integrate a frequency-based module to the algorithm. The motivation for this comes from the assumption that pixels that are probable count as chest pixels will demonstrate a periodic behaviour in their intensity depth value over time, as the patient inhale and exhale back and forth.
One of the ways to check for the periodicity of a signal is to break it to several periodic functions, by applying the Fourier Transform on the data over time.
Let $V$ be the volumetric video matrix of size (\(l\times m\times  n\)) where $m$ and $n$ are the image width and height and $l$ is the length of the depth sequence.
Let \(S_{nm}(t) \) be the pixel signal along in time $t$. We measure the signal along time interval segments \(\Delta t\). The algorithm will transform each of those signals to  \(F_s(\omega\)), representing the Fourier Transform decomposition of the pixel signal in the frequency domain. Each of the transforms represents a sequence of a single pixel over time.
After this transformation is calculated, the maximal amplitude that can be found is returned, representing the dominant amplitude of the signal (while still ignoring noisy pixels that were extracted before) creating an amplitude image for the specific sequence that corresponds to the peaks in frequency for each segment that can represent the level of periodicity estimated to each one of the pixels in the depth sequence.
An example of such image can be seen in Fig.~\ref{fig:MaximalAmplitude}.

The figure demonstrates that the chest area provides high amplitudes that are an indication of the periodic movement of the breathing pixels.

After evaluating the frequency matrix, a threshold is applied on the frequency image to include the relevant pixels that are part of the chest area – the most dominant periodic frequencies, as can be seen in Fig.~\ref{fig:MaximalAmplitude}.

It can be seen that the segmented pixels still contain some irrelevant noise pixels, a matter that will be covered in the next section, but generally, most of the chest area is segmented and marked.


\subsubsection{Noise in frequency domain}
\begin{figure}[!htb]
    \begin{floatrow}
    \subfloat[Noise pixel]{{\includegraphics[width=0.45\linewidth]{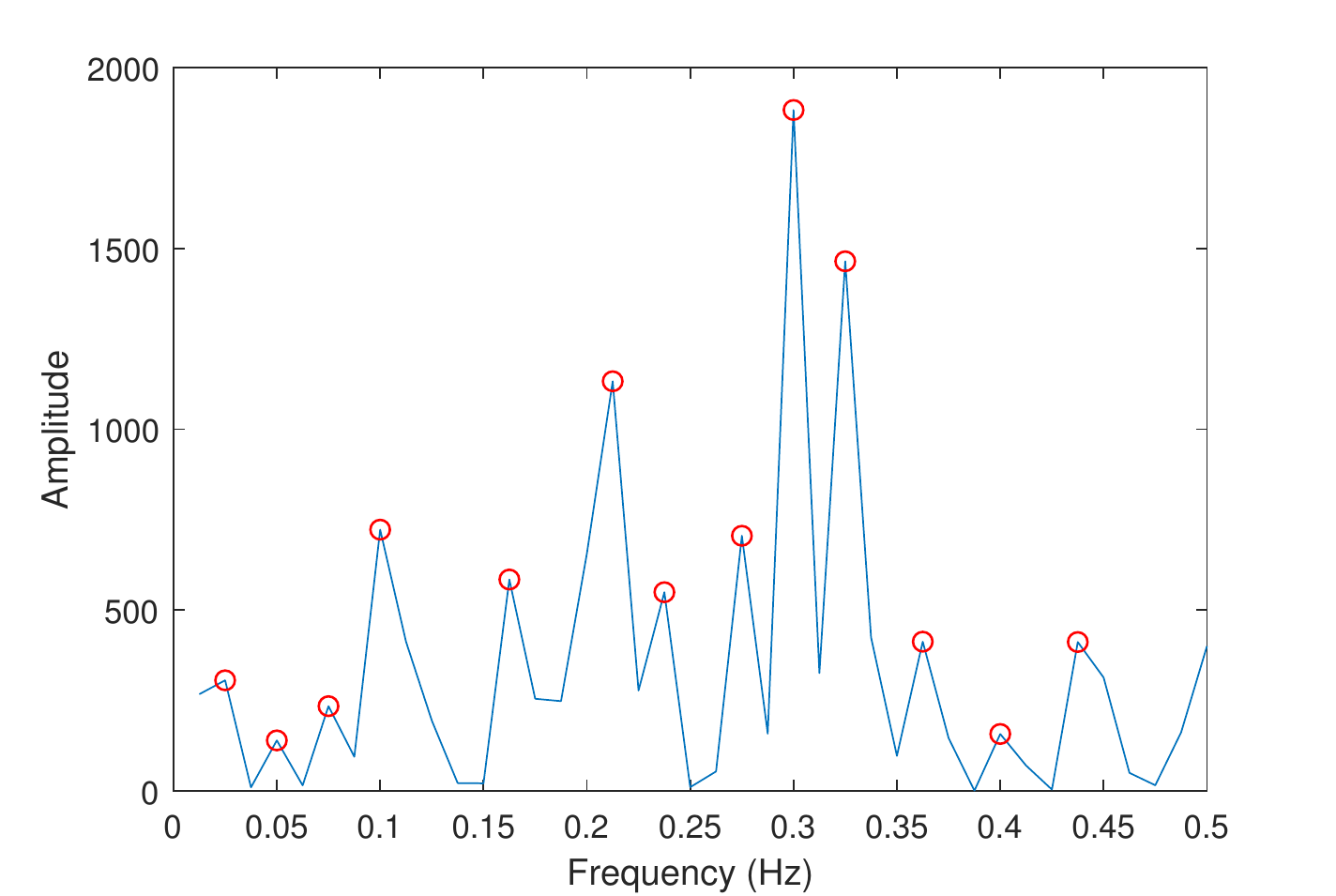} }}%
    \qquad
    \subfloat[Chest Pixel]{{\includegraphics[width=0.45\linewidth]{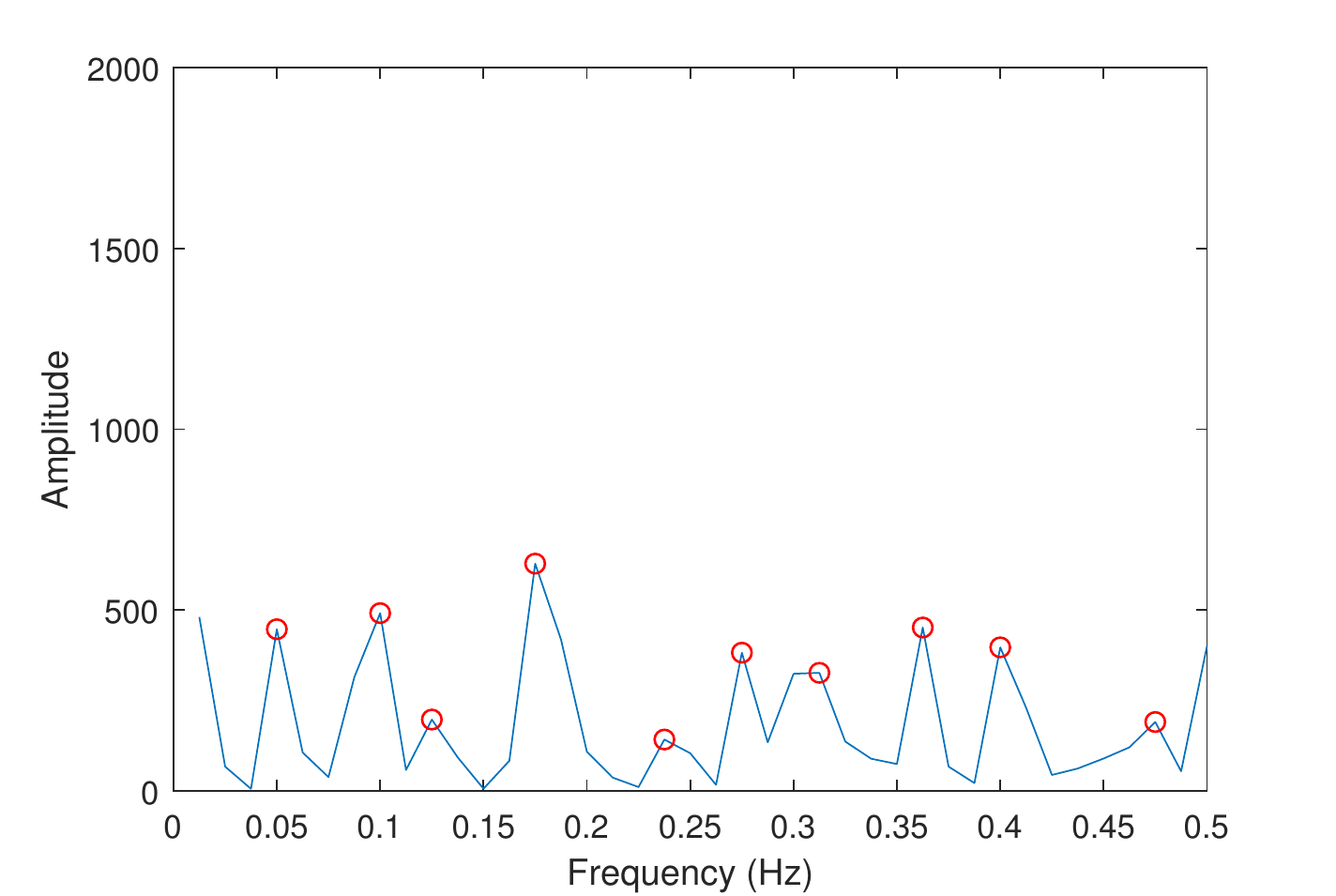} }}%
    \caption{Examples of two frequency peaks plots of noisy pixel (a) and chest pixels (b)}%
    \label{fig:SamplePixels}%
    \end{floatrow}
\end{figure}

Even though initial frequency domain manipulations on the pixels over time obtained segmentation of pixels that relate to the breathing, the depth data still remained very high in noise, allowing noise pixels produce high enough amplitudes that overcame the basic threshold and to be counted as a chest pixels. This occurred due to some level of periodicity that reflects in the high amplitude of some other frequencies that might not be related to the breathing.
As explained in \cite{Prochazka2016}, breathing frequency ranges from 12 to 20 bpm $(0.2Hz$ to $0.33Hz)$. Therefore, several pixels were selected from two groups - one represents noisy pixels that contained high peaks in frequency and second one with actual chest pixels. A frequency plot of those pixels can be seen in Fig.~\ref{fig:SamplePixels}.

\begin{figure}[!htb]
    \centering
    \begin{floatrow}
    \subfloat[Segmentation results after applying filter in the frequency domain, removing unnecessary high frequencies.]{{\includegraphics[width=4.5cm]{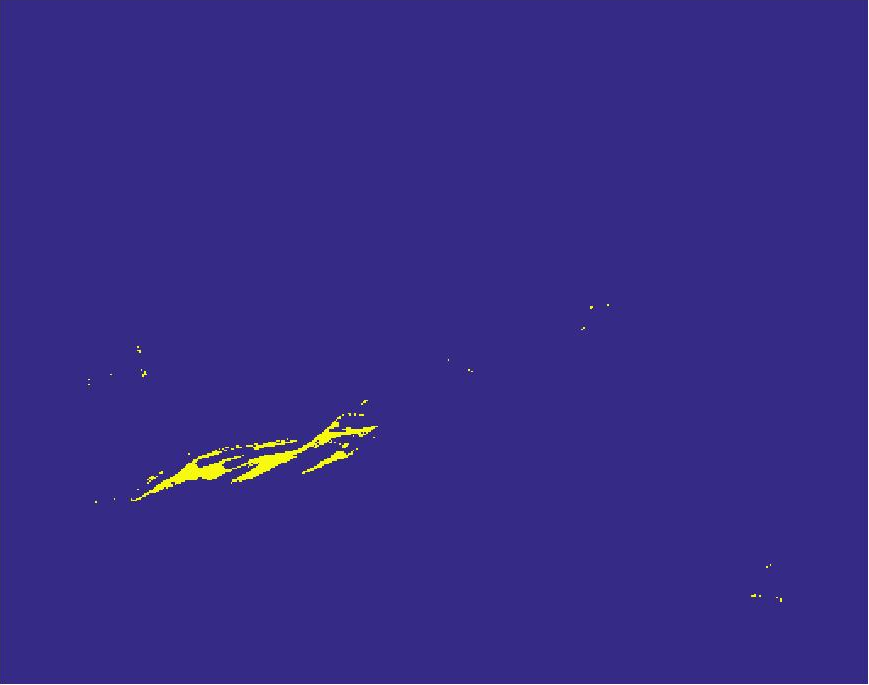} }}%
    \qquad
    \subfloat[The chest area segmentation for a single time segment after applying the final morphological operations.]{{\includegraphics[width=4.5cm]{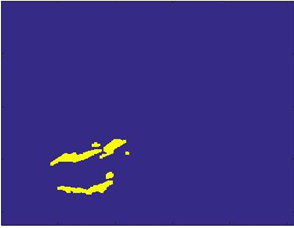} }}%
    \caption{Segmentation results after filter (a) and segmentation after morphological operations(b)}%
    \label{fig:SegmentationResultsAfterLPF}%
    \end{floatrow}
\end{figure}

As seen, in ~\ref{fig:SamplePixels}a clear peaks in the range of the breathing frequency are visible, while in the second example of noise pixel in ~\ref{fig:SamplePixels}b, the peaks are distributed throughout the entire range.
This led to an idea to discard all frequencies that lie outside the breathing frequency range, by applying a band-pass filter on the frequency domain. This lead to better segmentation results, as can be seen in Fig.~\ref{fig:SegmentationResultsAfterLPF}. As can be observed, the noisy irrelevant pixels had been reduced almost entirely, leaving almost only the relevant chest pixels.

As can be observed, the noisy irrelevant pixels had been reduced almost entirely, leaving almost only the relevant chest pixels.


\subsubsection{Morphological Operations}

As a final step, to eliminate the remaining segmentation noise that has not been successfully removed so far, morphological operations were applied on the binary image. These operations removed additional irrelevant pixels, leaving almost only chest related pixels, in addition to closing holes in the relevant sections.
An example of the output can be observed in Fig.~\ref{fig:SegmentationResultsAfterLPF}. As is shown, the chest area segmentation is clearly visible.


\subsection{Time Segmentation Algorithms}
This following section describes the variety of algorithm applied on the data over time in order to obtain more accurate segmentation map of the chest area.

\subsubsection{Accumulated Segmentation and Confidence Map}

\begin{figure}[!htb]
    \centering
    \begin{floatrow}
    \subfloat[Confidence map from the sequences over time, brighter pixel is more likely to be a chest pixel.]{{\includegraphics[width=5cm]{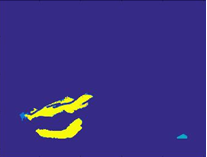} }}%
    \qquad
    \subfloat[Final Segmentation map of the chest pixels.]{{\includegraphics[width=5cm]{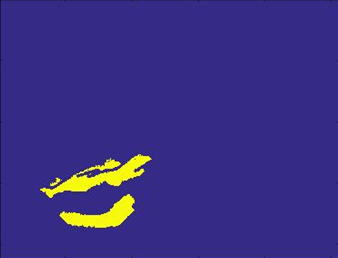} }}%
    \caption{Confidence map and final Segmentation map.}%
    \label{fig:ConfidenceMap}%
    \end{floatrow}
\end{figure}

Since the algorithm above was applied on time segments from the video, the data from each segment was accumulated to a histogram image in which every pixel that was identified as a chest pixel in a specific section increased the corresponding pixel in the histogram image by one, creating an accumulated image for the entire video sequence.
Later on, the image was normalized to values in the [0,1] range to be treated as a confidence  map (Fig~\ref{fig:ConfidenceMap}a). A bright pixel in the image will represent a pixel that segmented as a chest pixel many times and therefore the image can represent a confidence map of the pixel probability to be a chest pixel.
Applying another confidence threshold on the normalized image allowed to remove additional noise and to better segment the relevant pixels and remove outliers (Fig~\ref{fig:ConfidenceMap}b). An example of two confidence maps can be shown in Fig~\ref{fig:ConfidenceMap}.

\subsubsection{Temporal threshold by confidence}
A final threshold was applied on the confidence map in order to leave only pixels that are more probable to belong to the chest area, and an output binary image was created that represents the final segmentation for the entire video sequence. As was shown in \cite{Prochazka2016} this counts as a good starting point to evaluate the breathing signal and the sleep features required for a sleep study. The final segmentation can be shown in Fig.~\ref{fig:ConfidenceMap}b.


\section{Results}

\vspace{-0mm}

The algorithm results were evaluated according to three key parameters:
\begin{enumerate}
\item The ability of the algorithm to perform well on multiple sleeping poses.
\item The ability to perform on a variety of patients.
\item The ability of the algorithm to handle several levels of noise.
\end{enumerate}

To evaluate its performance, data from multiple sleeping patients was obtained from the research performed in \cite{Prochazka2016} and the algorithm was applied to it. 

\subsection*{\textbf{Multiple poses}}
As the algorithm looks for the periodic motion of specific pixels in the chest region, it should be invariant to the various sleeping poses, even if the chest area is not exposed to the camera, the result of the breathing movement still can be detected.

\begin{figure}
[!htb]
\begin{center}  
\includegraphics[width = 0.8\linewidth]{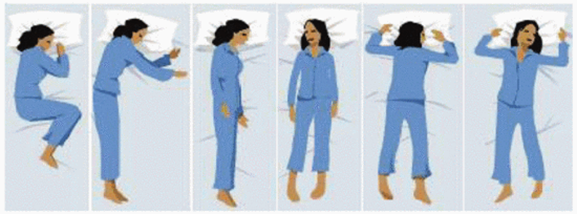}
\caption{\small \sl Most common sleeping postures for classification. From left to right are: 1. left-foetus, 2. left-learner, 3. left-log, 7. soldier, 8. free-fall, 9. star-fish. (4.-6. are right-lying postures of 1.-3.) \cite{Huang2010}.
\label{fig:sleepingPosture}}  
\end{center}  
\end{figure}

The main sleep poses can be seen in Fig.~\ref{fig:sleepingPosture}.

\begin{figure}
[!htb]
    \centering
    \subfloat[Left-Log sleeping posture ]{{\includegraphics[width=0.45\linewidth]{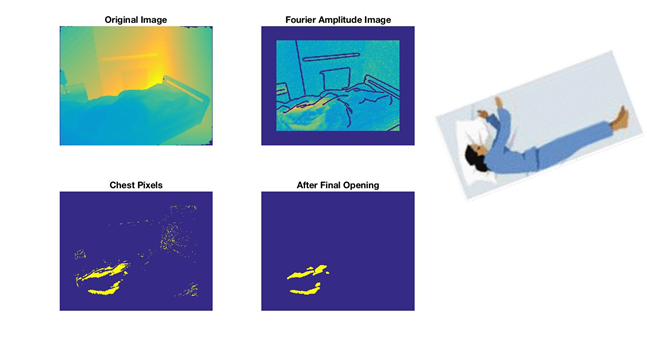} }}%
    \subfloat[soldier sleeping posture]{{\includegraphics[width=0.45\linewidth]{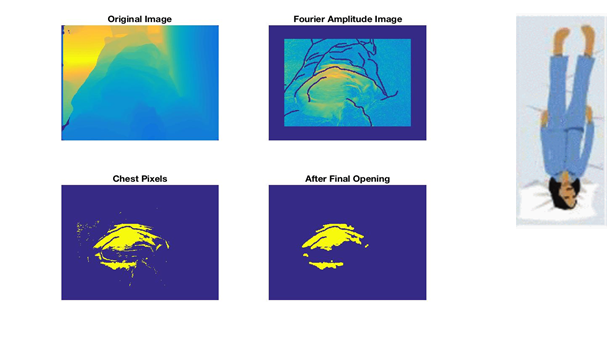} }}%
    \caption{Comparison of the algorithm output between  patients with various sleeping positions}%
    \label{fig:SleepingPoses}%
\end{figure}

As the data is recorded in a sleep clinic while the patient is connected to several sensors, there are no recordings of patients sleeping in ‘free fall’ position since this posture interferes with the PSG sensors. The algorithm was tested to handle patients with multiple sleeping poses, also with multiple angles (from the side of the bed, from the top of the bed). Some of the segmentation results are shown in Figure
\ref{fig:SleepingPoses} which shows invariance in an example of two sleeping positions.





\subsection{\textbf{Multiple Patients}}
\begin{figure}
[!htb]
    \centering
    \subfloat[Patient 1]{{\includegraphics[width=0.45\linewidth]{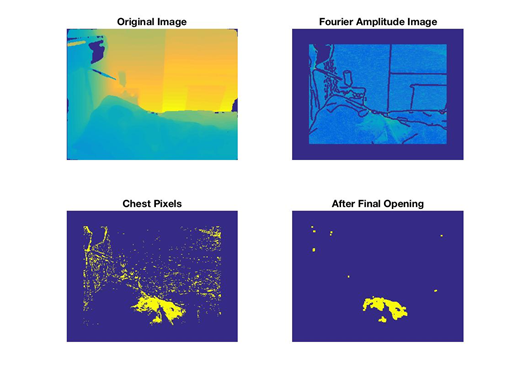} }}%
    \subfloat[Patient 2]{{\includegraphics[width=0.45\linewidth]{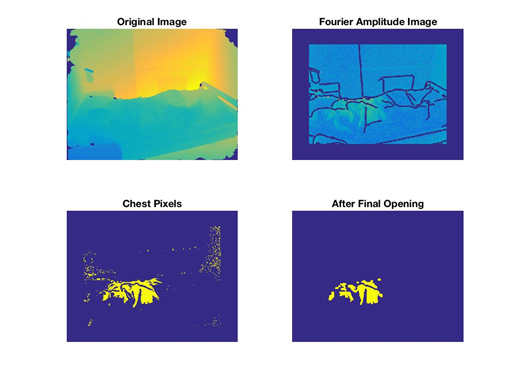} }}%
    \caption{Comparison of the algorithm output between various patients with the same sleeping position}%
    \label{fig:ComparisonSamePosition}%
\end{figure}

The algorithm also operated on same patient captured during same posture, as shown in Fig.~\ref{fig:ComparisonSamePosition}.

Although different patients have different sleeping cycles and their breathing patterns are different, the segmentation output can be invariant to different kinds of patients. A possible explanation for this ability can arise from the periodic behaviour of every patient sleep movement and the ability of the algorithm to recognize it and utilize it for the segmentation.

\subsection{\textbf{Noise Reduction}}

Noise from the depth has been significantly reduced using both spatial and temporal operations. The multiple noise reduction pipeline performed to obtain the maximal number of relevant chest pixels while minimizing noise on the segmentation output.
Fig.~\ref{fig:NoiseReductionResult} demonstrates the noise reduction results:

\begin{figure}[H]
\begin{center}  
\includegraphics[width = 0.5\linewidth]{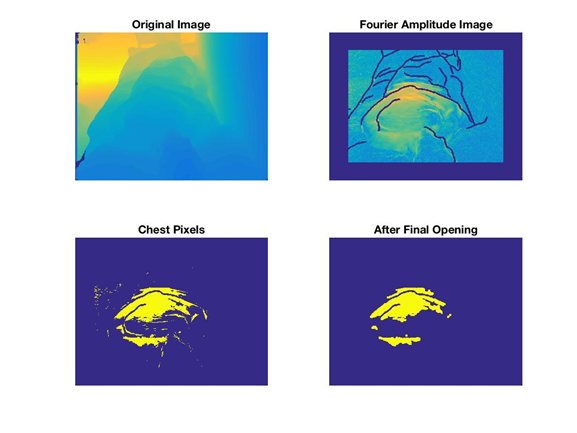}
\caption{\small \sl  original image after median filter (top left), frequency amplitude image with edges and frame noise removed (top right), chest pixel after segmentation and noise reduction in frequency domain (bottom left), final segmentation after morphological operations (bottom right).
.\label{fig:NoiseReductionResult}}  
\end{center}  
\end{figure}

\begin{figure}[H]
\begin{center}  
\includegraphics[width = 0.9\linewidth]{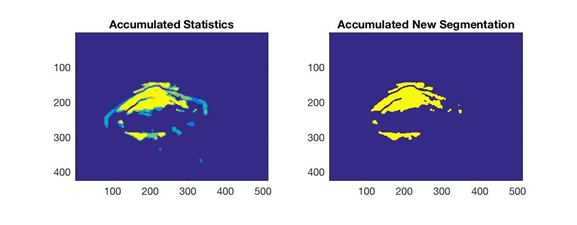}
\caption{\small \sl  The confidence map of the entire frame sequence (left) and the final segmentation for the entire movie after confidence thresholding (right).
.\label{fig:FinalSegmentationResult}}  
\end{center}  
\end{figure}

As can be seen in the figure,  only small amount of noise that is a result of noise in the current processed video segment is left. However, it is treated over time, as shown in Fig.~\ref{fig:FinalSegmentationResult}.

As explained earlier, the confidence map represents the certainty of pixels to belong to the chest along time. Indeed, after time threshold, it can be seen that some of the noise that remained is indeed removed, leaving the chest pixels segmentation more accurate.


\subsection{Accuracy Enhancement}
\begin{figure}
[!htb]
\begin{center}  
\includegraphics[width = 0.9\linewidth]{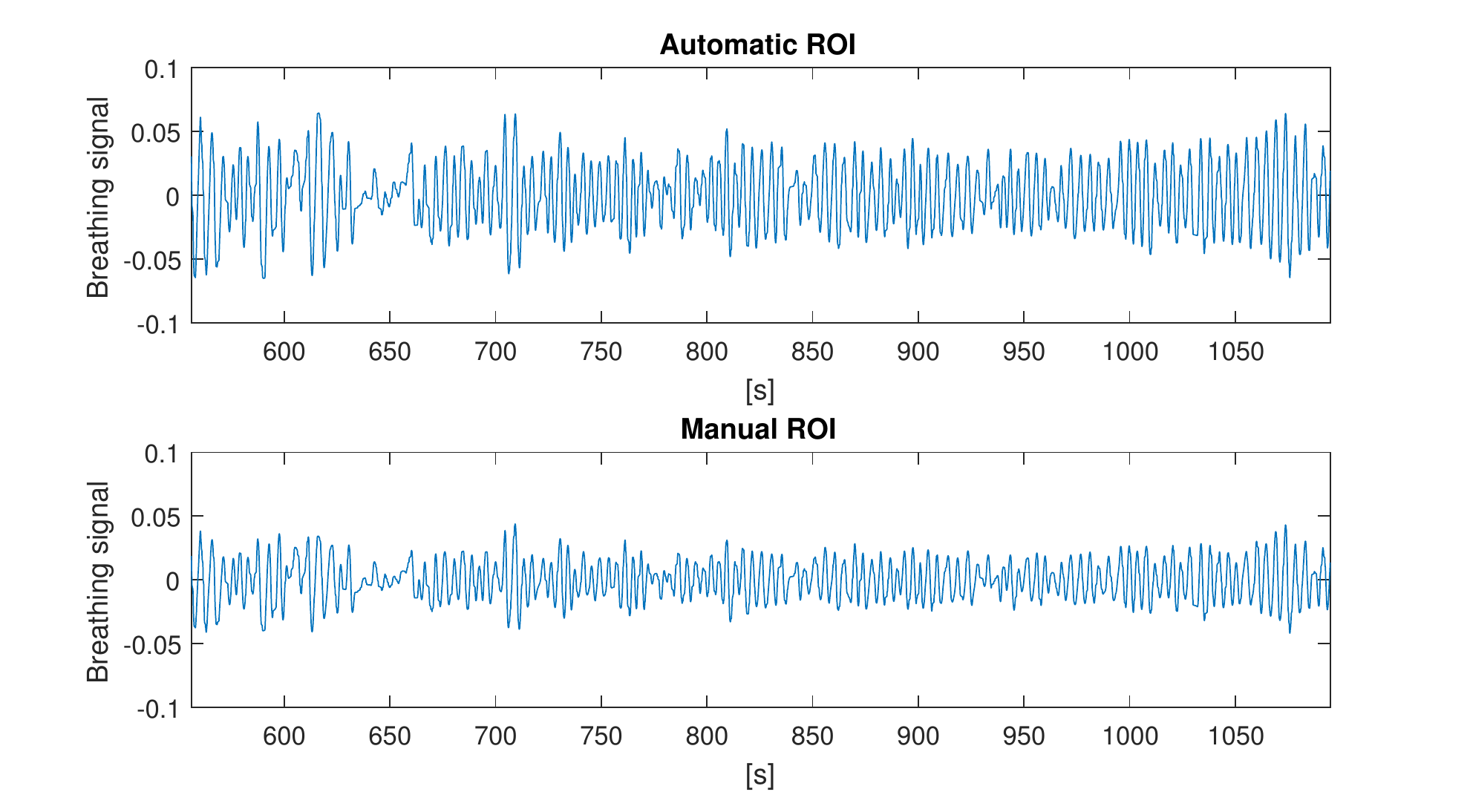}
\caption{\small \sl  Original breathing signal extracted from the depth sequence (top) compared to the signal extracted using the segmentation algorithm (bottom).
.\label{fig:CompResult}}  
\end{center}  
\end{figure}

Obtaining the segmentation map contributed to further develop the breathing frequency extraction algorithm described in \cite{Prochazka2016}. However it has remained to wonder weather indeed its addition contributed to its accuracy. Therefore, the segmentation algorithm was integrated to \cite{Prochazka2016}. A comparison between the two breathing signals, with and without the segmentation is visible in Fig \ref{fig:CompResult}.

\begin{figure}
[!htb]
    \centering
    \subfloat[Features on original signal]{{\includegraphics[width=0.45\linewidth]{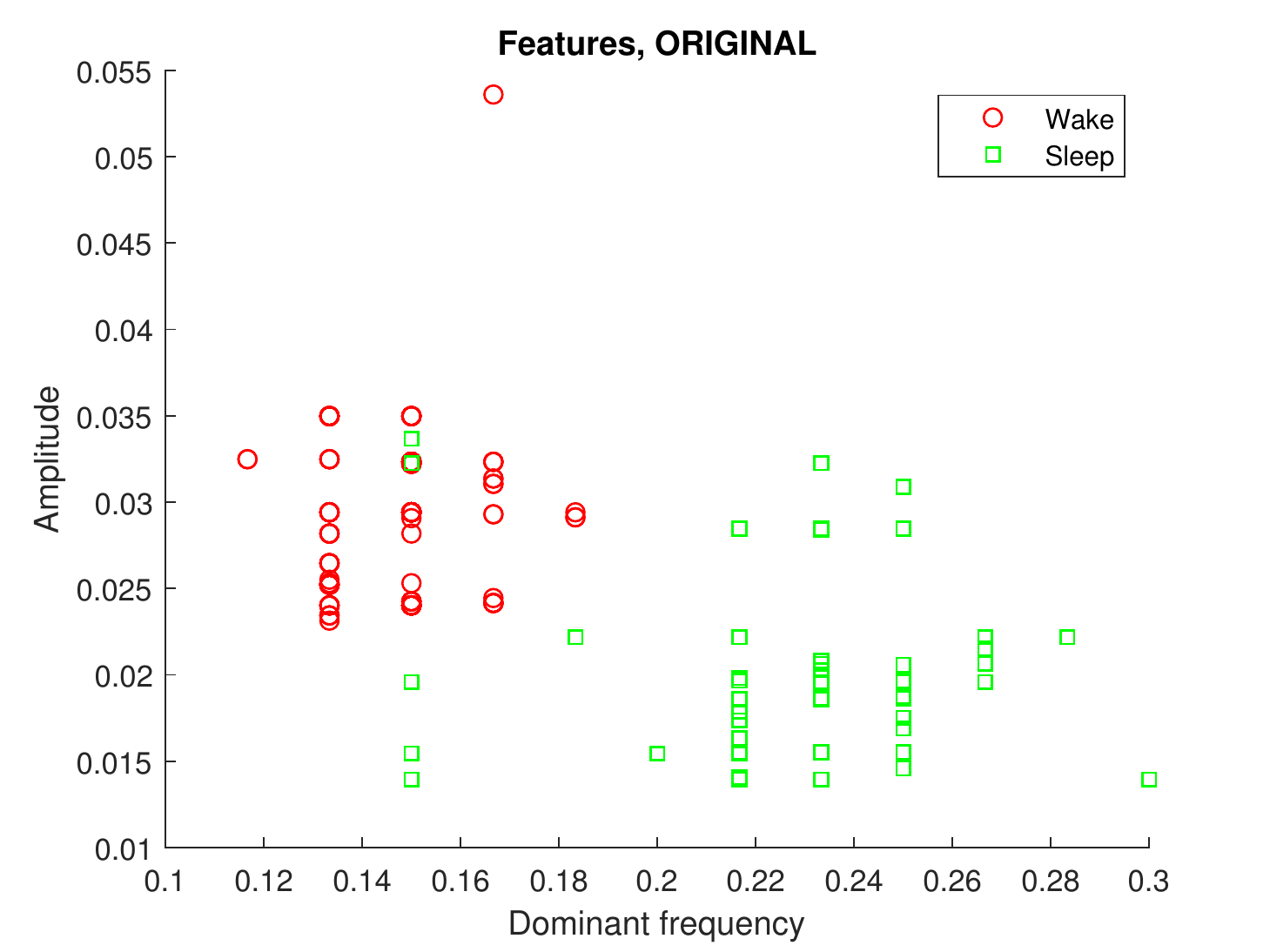} }}%
    \subfloat[Features using a signal created using the segmentation map]{{\includegraphics[width=0.45\linewidth]{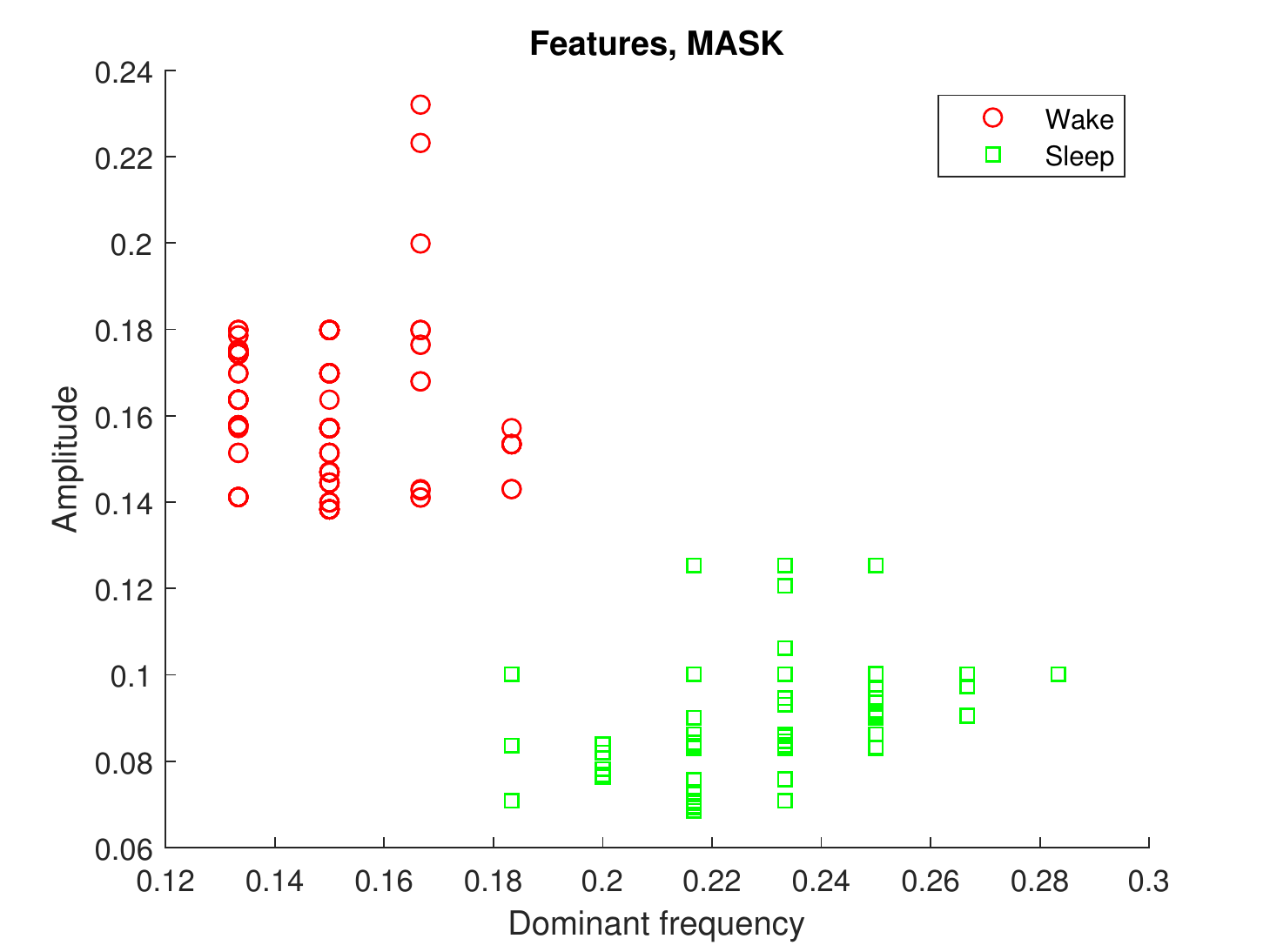} }}%
    \caption{- Classified features of the original breathing signal (left) and of the breathing signal obtained by segmentation (right)}%
    \label{fig:FeaturesComparison}%
\end{figure}

The data was recorded using signals from Kinect depth sensor (recorded from 1.5 m) v.s. spectrum of a signal from Polysomnography. The comparison of manual v.s. automatic ROI detection (with the segmentation algorithm), is on average 46.9\% more sensitive with a maximal improvement of 220\% when compared to manual ROI.
The features of the breathing signals were inserted into the classifier and each feature, representing a sub sequence in the breathing signal, was given a label, classifying it to wake or sleep. The result of this classification, on both breathing signals, can be seen in Fig \ref{fig:FeaturesComparison}

It can be seen that that on the same signal, a much clearer separation between the sleep and wake modes can be achieved using the segmentation pixels from the breathing area.




\vspace{3mm}

\section{Conclusions}
\vspace{-0mm}

We showed that it is feasible to perform automatic chest area segmentation on input depth images of sleeping patients.
As stated in \cite{Prochazka2016}, the chest region in depth images stream is the base for the extraction of the breathing signal which is later used to evaluate the breathing frequency and other sleeping parameters. The research in \cite{Prochazka2016} used a rough rectangle manually selected to include the patient chest area for the algorithm. By achieving the chest individual pixel segmentation, the ROI is no longer needed to be manually selected and is now automatically recognized. This can allow much more precision in the sleep feature extraction and research. In addition, while this algorithm achieved good results performing on the data, it was only tested on it and future work could be needed to allow it to perform on additional datasets and eventually generalize to any data.

\vspace{5mm}

\section*{Conflicts of Interest} \vspace{-0mm}

The authors declare no conflict of interest.

\vspace{3mm}

\section*{Acknowledgment}
Real data were kindly provided by the Department
of Neurology of the Faculty hospital of the Charles University in
Hradec Kralove. The project was approved by the Local Ethics
Committee as stipulated by the Helsinki Declaration.


\bibliographystyle{IEEEtran}
\bibliography{ Chest_Area_Segmentation_in_Depth_Images_of_Sleeping_Patients, MS, dsp16}

\end{document}